\documentclass{article}
\usepackage{spconf,amsmath,graphicx}
\usepackage[ruled]{algorithm2e}
\usepackage{caption}
\usepackage{subcaption}

\DeclareMathOperator*{\argmax}{argmax}

\title{Deep Flow Collaborative Network For Online Visual Tracking}
\name{Peidong Liu \qquad Xiyu Yan \qquad Yong Jiang \qquad Shu-Tao Xia}

			\address{Tsinghua Shenzhen International Graduate School, Tsinghua University, China \\
			  	\{lpd19, yanqy17\}@mails.tsinghua.edu.cn;
				\{jiangy, xiast\}@sz.tsinghua.edu.cn}

\begin{document}
%
\maketitle
\begin{abstract}
The deep learning-based visual tracking algorithms such as MDNet achieve high performance leveraging to the feature extraction ability of a deep neural network. However, the tracking efficiency of these trackers is not very high due to the slow feature extraction for each frame in a video. In this paper, we propose an effective tracking algorithm to alleviate the time-consuming problem. Specifically, we design a deep flow collaborative network, which executes the expensive feature network only on sparse keyframes and transfers the feature maps to other frames via optical flow. Moreover, we raise an effective adaptive keyframe scheduling mechanism to select the most appropriate keyframe. We evaluate the proposed approach on large-scale datasets: OTB2013 and OTB2015. The experiment results show that our algorithm achieves considerable speedup and high precision as well.
\end{abstract}
\section{Introduction}
\label{sec:intro}

Visual tracking task\cite{visual_tracking_1, visual_tracking_2} has attracted significant attention from researchers due to a wide range of potential applications such as VR, traffic control, robots, surveillance systems, etc. However, it is a challenging task because of the environmental variation,  appearance variation of the target, and the high-efficiency requirements for some applications. An excellent tracker should consider both model robustness and effectiveness. For one thing, model robustness means the tracker performs well even when in complex environments such as background clutter, illumination variation, etc. Besides, the tracker can adapt to the appearance variation of the tracking object. For another thing, model effectiveness indicates the tracker achieves adequate speed for the applications.

With the development of the convolution neural network (CNN) \cite{krizhevsky2012imagenet,simonyan2014very,Szegedy_2015_CVPR,he2016deep}, deep learning-based methods are explored broadly in the visual tracking area. Benefitting from better feature representation extracted\cite{profile_1, profile_2, texture_1, texture_2} by CNN, deep learning-based trackers such as MDNet\cite{MDNET}, HDT\cite{HDT}, and SINT\cite{SINT+} have obtained high accuracy in the large-scale benchmark, which means that they have excellent model robustness. However, most of these trackers haven't taken model effectiveness into full consideration. Specifically, they are time-consuming due to the complex architectures and large numbers of computations during feature extraction. Some 
previous works such as Real-time MDNet\cite{realtime_mdnet} and FlowTrack\cite{flowtrack} have explored the model effectiveness. However, Real-time MDNet\cite{realtime_mdnet} realizes the model effectiveness at the cost of decreasing the model robustness. While FlowTrack\cite{flowtrack} needs to spend time training the whole model, including flow sub-network and feature sub-network, before the tracking process, which requires lots of computations. 

\begin{figure}[!t]
\centering
\includegraphics[width=0.45\textwidth]{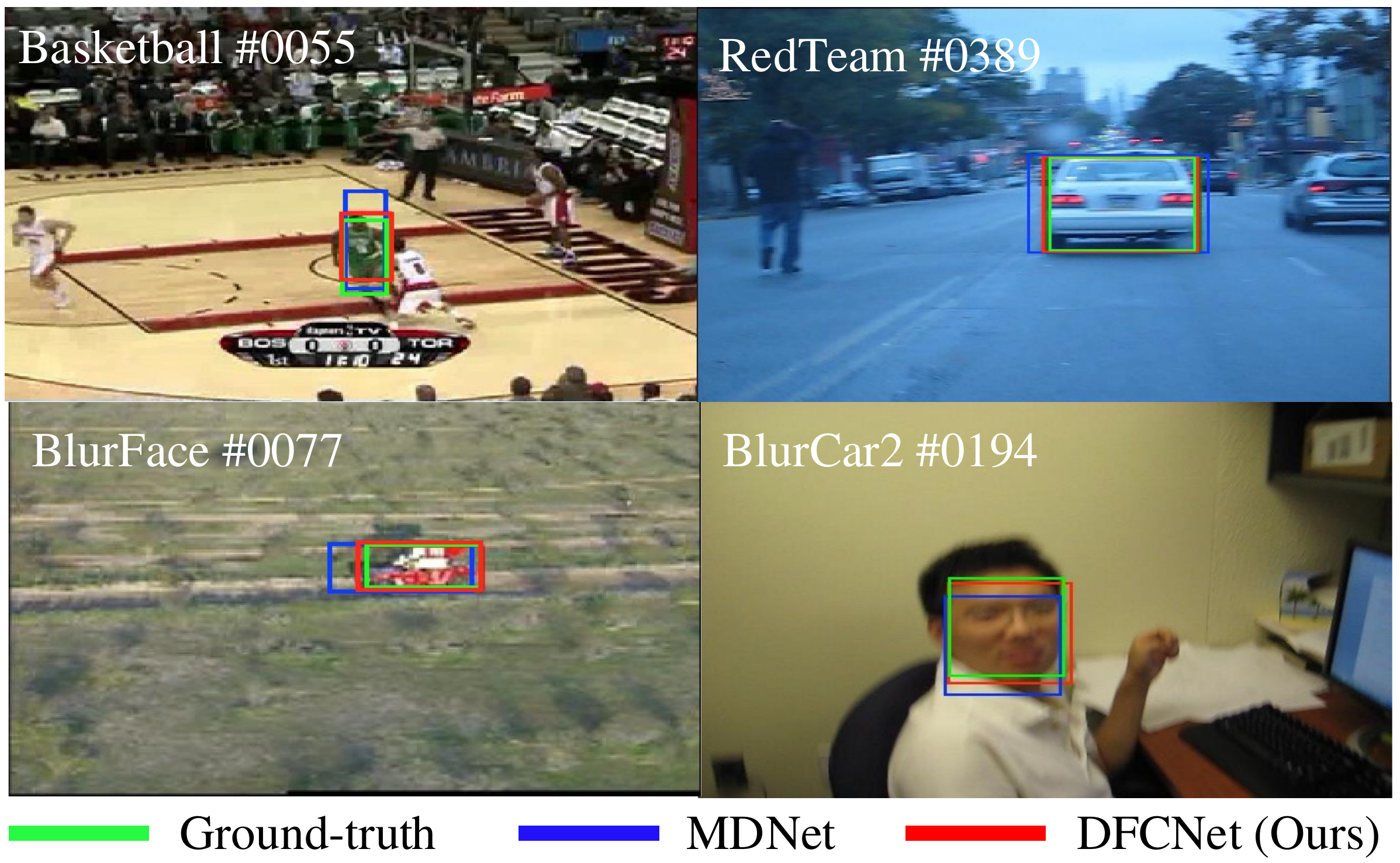}
\caption{Tracking results comparison of our approach DFCNet with MDNet\cite{MDNET} in the challenging scenarios.}
\label{fig:comare_pic}
\end{figure}


\begin{figure*}[!t]
\centering
\includegraphics[width=0.8\textwidth]{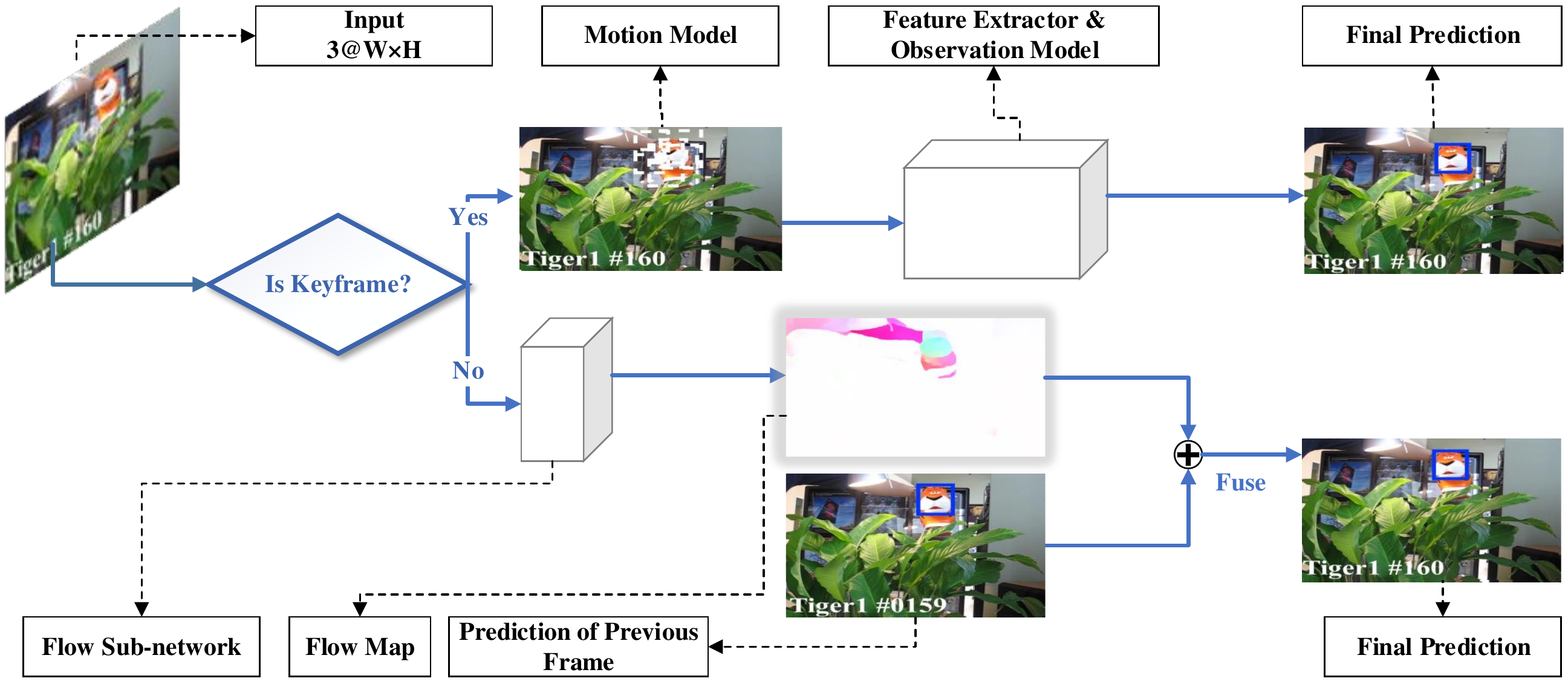}
\caption{The illustration of DFCNet Architecture, including the keyframe branch (motion model, feature extractor, and observation model) and non-keyframe branch (flow sub-network).}
\label{fig:architecture_DFCNet}
\end{figure*}

In this work, we aim to propose an effective online algorithm that can fully consider model effectiveness to alleviate the time-consuming problem under the premise of maintaining model robustness at the same time. To be specific, we design a deep flow collaborative network (DFCNet) which utilizes inter-flow information in consecutive video sequences. As applying complex feature extractor to each frame is expensive, we speed up the tracking process by running the feature network only on sparse keyframes while other target states can be propagated through an optical flow map. Besides, we propose an effective keyframe scheduling mechanism to utilize appearance representation and temporal information. Fig. \ref{fig:comare_pic} shows DFCNet maintains robustness in complex scenes including background clutter, illumination variation, in-plane rotation, and out-of-plane rotation.

  The contributions of this paper can be summarized as follows:
 \begin{itemize}
\item We propose a deep flow collaborative tracking algorithm to alleviate the low-efficiency problem. Besides, an effective adaptive keyframe scheduling algorithm is developed, which can help propagate the flow map efficiently and fully utilize both flow information and appearance feature.
\item In the experiments of OTB2013\cite{OTB2013} and OTB2015\cite{OTB2015}, the proposed algorithm is about 60\% faster than baseline model MDNet\cite{MDNET} while maintains model robustness at the same time. Our tracker performs favorably against most of the existing popular trackers in terms of robustness.
 \end{itemize}

\section{Proposed Method}
\label{sec:pagestyle}





















\subsection{Network architecture}
	Our proposed model extends the tracker MDNet\cite{MDNET}, champion of VOT2015\cite{VOT2015}, to the sub-network for feature extraction. Yet, the average speed of the MDNet\cite{MDNET} is around 1.55 fps\cite{visual_tracking_2} due to the generation of a large number of candidate samples and then the feature extraction through the deep neural network. 

	To speed up, DFCNet adopts a fast and effective method to avoid redundant computations and get accurate results by introducing optical flow. Only the sparse keyframes run the expensive feature network, and other target states are obtained through optical flow calculated with previous frames. As the difference between adjacent frames is limited, temporal information can be gained by optical flow. The specific network architecture is shown in Fig. \ref{fig:architecture_DFCNet}. DFCNet first determines whether the current one is a keyframe. On the one hand, if it is exactly a keyframe, then we first generate a large number of candidate samples with the motion model, then extract features through the complex feature network, and finally obtain the estimated target with the observation model. On the other hand, if it is not a keyframe, we get the optical flow between the current frame and the previous frame through FlowNet2\cite{FLOWNET2} and then integrate the estimated result of the previous frame with the corresponding optical flow to determine the current target.

\subsection{Adaptive keyframe scheduling mechanism}
	As for DFCNet, only keyframes pass through the feature extraction network. If all frames are determined to be keyframes, DFCNet degrades to the MDNet\cite{MDNET}. If only the first frame is a keyframe and the remaining are non-keyframes, due to the limited modeling of the appearance features, the tracker gets a poor accuracy. So it shows the significance of the keyframe selection strategy. Therefore, we propose a novel adaptive keyframe scheduling algorithm to get both speedup and high accuracy.

	To begin with, DFCNet sets an interval $K$ for scheduling mechanism to determine the fixed keyframes in a video sequence. For example, say $K$ = 3, frames such as $1^{st}$, $4^{th}$, $7^{th}$ and so on are considered as keyframes. The remaining frames are judged whether to be a keyframe or not in real-time based on the current tracking result. On the one hand, if the online tracking model scores higher than a given score threshold $T$, it indicates that the current tracking result is relatively accurate. Then the estimated optical flow is used to obtain the inter-frame information to complete the tracking process of the next frame. On the other hand, if the online tracking process has poor performance, the target of the next frame needs to be obtained through the feature network, which requires a lot more computations.

\subsection{Online tracking algorithm}
	DFCNet is an online tracking algorithm. Online tracking refers to fine-tuning the tracker online with the addition of the sequences. In detail, DFCNet adopts the pre-trained tracking model in the beginning, and then the model is updated in the subsequent frames. Online tracking is suitable for the sequences as it can provide more accurate predictions with the increase of the input data. 

	DFCNet can be divided into several modules, including motion model, feature extractor, observation model, and model update. The online tracking algorithm is presented in Algorithm \ref{algo:flame}. The detailed tracking procedure is discussed in the following. 

\begin{algorithm}[!t]
\KwIn{Initial target state $B_0$, Number of frames in a video $C$, Number of point samples $M$, Score threshold $T$}

\KwOut{Estimated target state $\{B_i\}_{i=1}^{N}$}

Initialize Score = +$\infty$.

\For{$i=1$ \emph{\KwTo} $C$} {
  \eIf {$i$ {\bf mod} $K == 1$ {\bf or} $Score <= T$} {
  \tcc*[f]{Keyframe}\qquad\qquad\qquad\qquad\ \ \ \ \ \ \ \ \
  
    $D = N_{Feature}(B_{i-1})$.\\
	Calculate $B_i$ according to Equation \ref{equ:argmax_feature}. \\
    $Score = f^+(B_i)$.\\    
  }
  {
	\tcc*[f]{Non-keyframe}\qquad\qquad\qquad\ \ \ \ \ \ \ 

    $F_{i-1 \rightarrow i} = N_{Flow}(I_{i-1}, I_i)$. 

    Sample $M$ points $\{P_j\}_{j=1}^M$ from $B_{i-1}$.

	Propagate $\{P_j\}_{j=1}^M$ to $\{P'_j\}_{j=1}^M$ via flow field.
  
	Adjust $\{P'_j\}_{j=1}^M$ slightly to get $B_i$.
  }
}

\caption{Deep flow collaborative algorithm for online visual tracking.}
\label{algo:flame}
\end{algorithm}

\subsubsection{Motion model}
	Motion model adopts particle filter\cite{particle_filter}. Based on the estimated target of the previous frame, the motion model generates $Q$ candidate bounding boxes following Gaussian distribution, which may contain the target of the current frame.

	In comparison, non-keyframes prevent running the motion model and generating large numbers of candidates.

\subsubsection{Feature extractor and observation model}
	The feature extractor converts the raw RGB image into a semantic feature representation. It is the most critical part of a tracker as informative features can boost the tracking result significantly. Then the observation model judges whether the candidate is the target based on the features extracted.

	In keyframes, DFCNet adopts a modified MDNet\cite{MDNET}, to extract features by replacing the last multi-domain layer with a single-domain layer. The modified network is composed of three convolutional layers (Conv1-3) and three fully connected layers (FC4-6). During the online tracking process, Conv1-3 and FC4-5 layers use pre-trained parameters to initialize, and only three fully connected layers are updated. After the feature sub-network, $N_{Feature}$ evaluates $Q$ candidate bounding boxes and obtains feature set $D$ of those candidates, DFCNet chooses the one with the highest positive score to be the estimated target as equation \ref{equ:argmax_feature} shows:
\begin{equation}
d^* = \argmax_{d} f^+(d), 
\label{equ:argmax_feature}
\end{equation}
where $d \in D$, $f^+$ is the positive score function, and $d^*$ is the optimal candidate sample. While in the remaining non-keyframes, estimated targets are obtained by fusing the target state of the previous frame and optical flow, which is obtained from flow sub-network $N_{Flow}$. The overall procedure of DFCNet is presented in Algorithm \ref{algo:flame}. Further details are described below.



\noindent \textbf{Shifting pixels in target state} DFCNet estimates the target state of non-keyframes through optical flow obtained from FlowNet2\cite{FLOWNET2}. Specifically, DFCNet gets estimated flow from $i^{th}$ to $i+1^{th}$ frame $F_{i \rightarrow i+1}$ and uniformly samples M pixels in the flow map to propagate the target from $i^{th}$ to $i+1^{th}$ frame.

\noindent \textbf{Bilinear interpolation in propagation} As the coordinate of pixels propagated may be floating-point numbers, we apply the bilinear interpolation to get the optical flow values in the flow map. In particular, the pixels in the target state of $i^{th}$ frame 
 \begin{equation}
\{(x_i^k,y_i^k) | k = 1,2,3,...,M \}
\end{equation}
can propagate to $i+1^{th}$ frame
\begin{equation}
	\begin{split}
		\{(&x_i^k+BIL(F_{i \rightarrow i+1}(x_i^k,y_i^k))[x], \\
		 &y_i^k+BIL(F_{i \rightarrow i+1}(x_i^k,y_i^k))[y]) | k = 1,2,3,...,M\},
	\end{split}
\end{equation}
where $BIL$ represents bilinear interpolation.

\noindent \textbf{Outliers removal and magnitude adjustment} Propagation may bring in some outliers, which result in poor accuracy. DFCNet keeps only $KR$ percent of propagated pixels to concentrate on the tracking object. As the appearance of objects in adjacent frames is similar, Hyperparameters adaptive ratio ($AR$) is used to balance the current and the previous target state to improve the robustness of the results.

\subsubsection{Model update}
	The model updating strategy refers to that of MDNet\cite{MDNET}, which mainly updates feature extractor. To fully consider model robustness and effectiveness, DFCNet applies long-term updates and short-term updates only on keyframes.

\section{Experiments}
\label{sec:majhead}

\subsection{Settings}
	We evaluated DFCNet on OTB2013\cite{OTB2013} and OTB2015\cite{OTB2015}. The feature sub-network is pre-trained on VOT2015\cite{VOT2015}, which excludes video sequences in OTB2015\cite{OTB2015}. The flow sub-network is pre-trained for video recognition. In online tracking, adaptive keyframe interval $K$ and score threshold $T$ is set to 3 and 10, numbers of candidate samples $Q$ and pixels in bounding box $M$ is set to 256 and 100, the ratio of reserved pixels $KR$ and the adaptive ratio $AR$ is set to 0.9 and 0.4, respectively. The one-pass evaluation (OPE) is applied to compare DFCNet with other trackers. For a fair comparison, all the tracking results use the reported results. Our algorithm is implemented in Pytorch and runs at a PC with 2.2GHz CPU and GTX1080 GPU.

\begin{figure}[!ht]
\begin{subfigure}{0.5\textwidth}
\centering
	\includegraphics[width=0.8\textwidth]{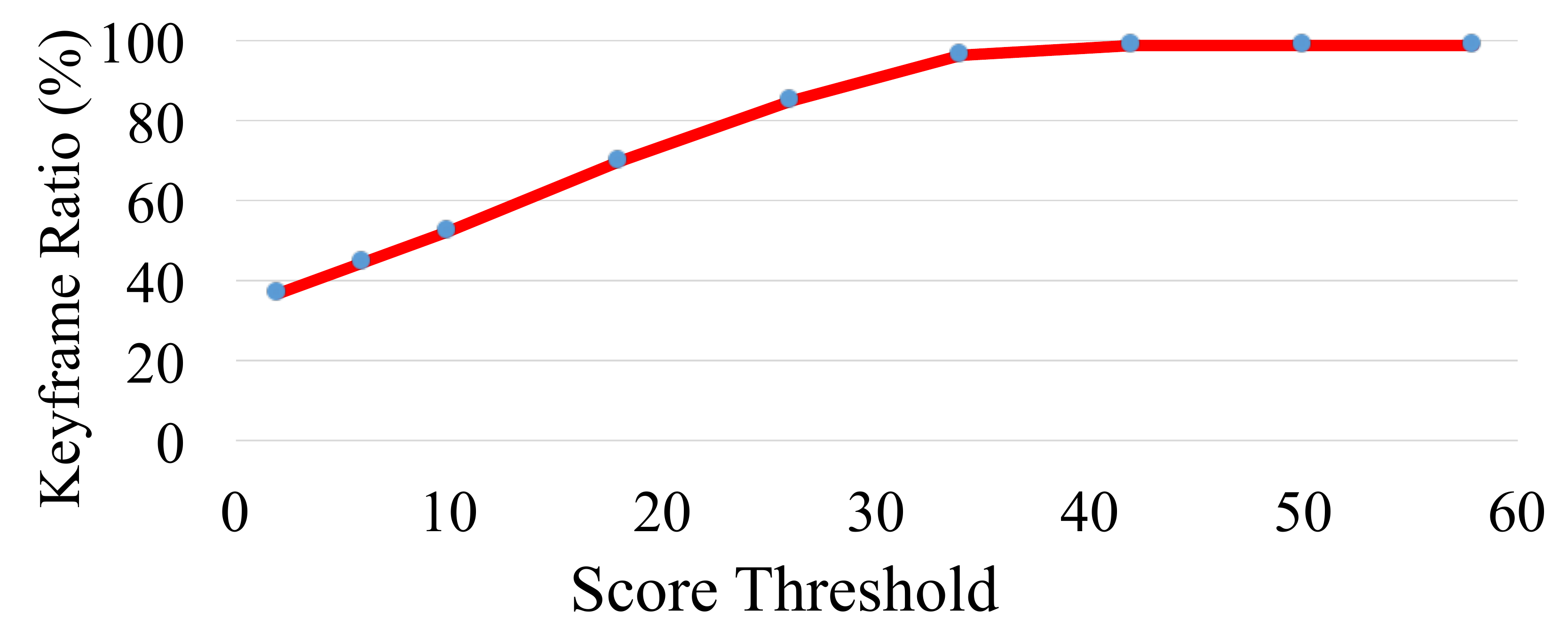}
\end{subfigure}
\begin{subfigure}{0.5\textwidth}
\centering
	\includegraphics[width=0.8\textwidth]{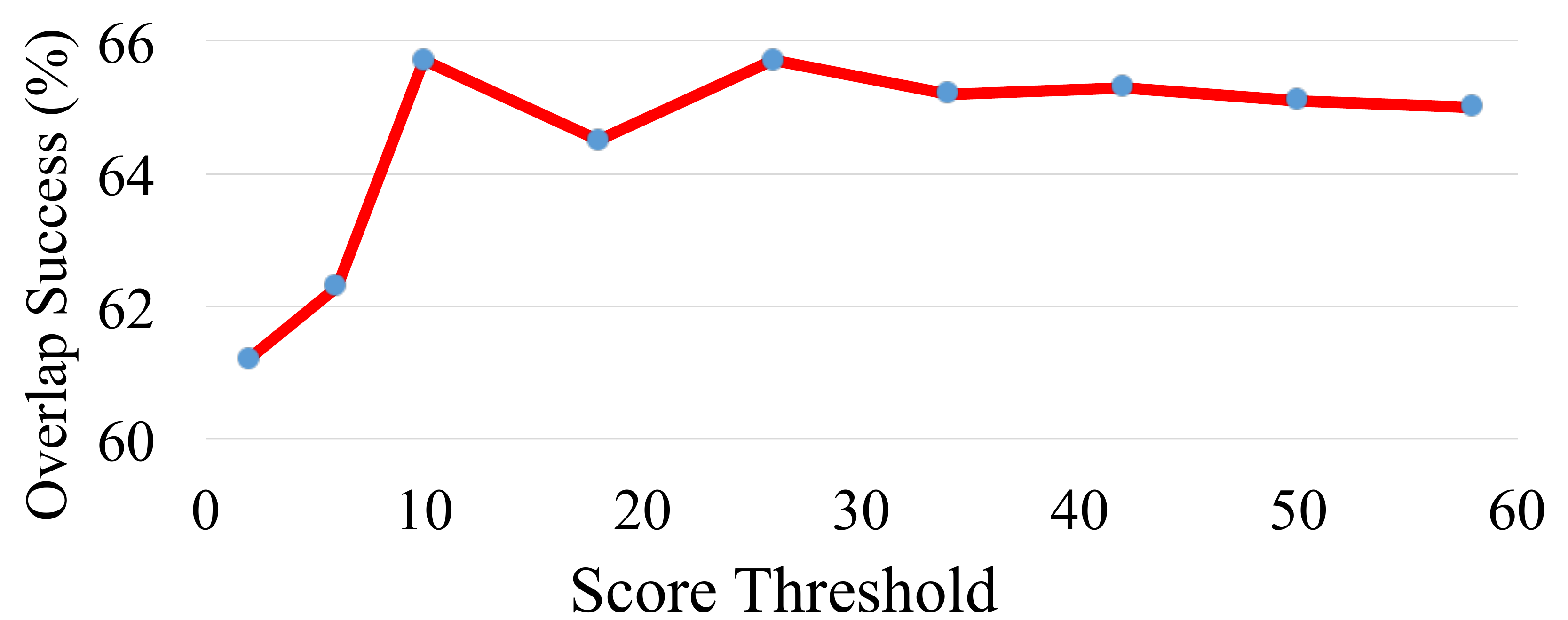}
\end{subfigure}
\caption{The above and below illustrations indicate the the relationship between keyframe ratio and score threshold, overlap success and score threshold, respectively.}
\label{fig:score_threshold_relation}
\end{figure}


\begin{figure}[!ht]
\centering
\begin{subfigure}{0.6\textwidth}
	\includegraphics[width=0.4\textwidth,height=0.35\textwidth]{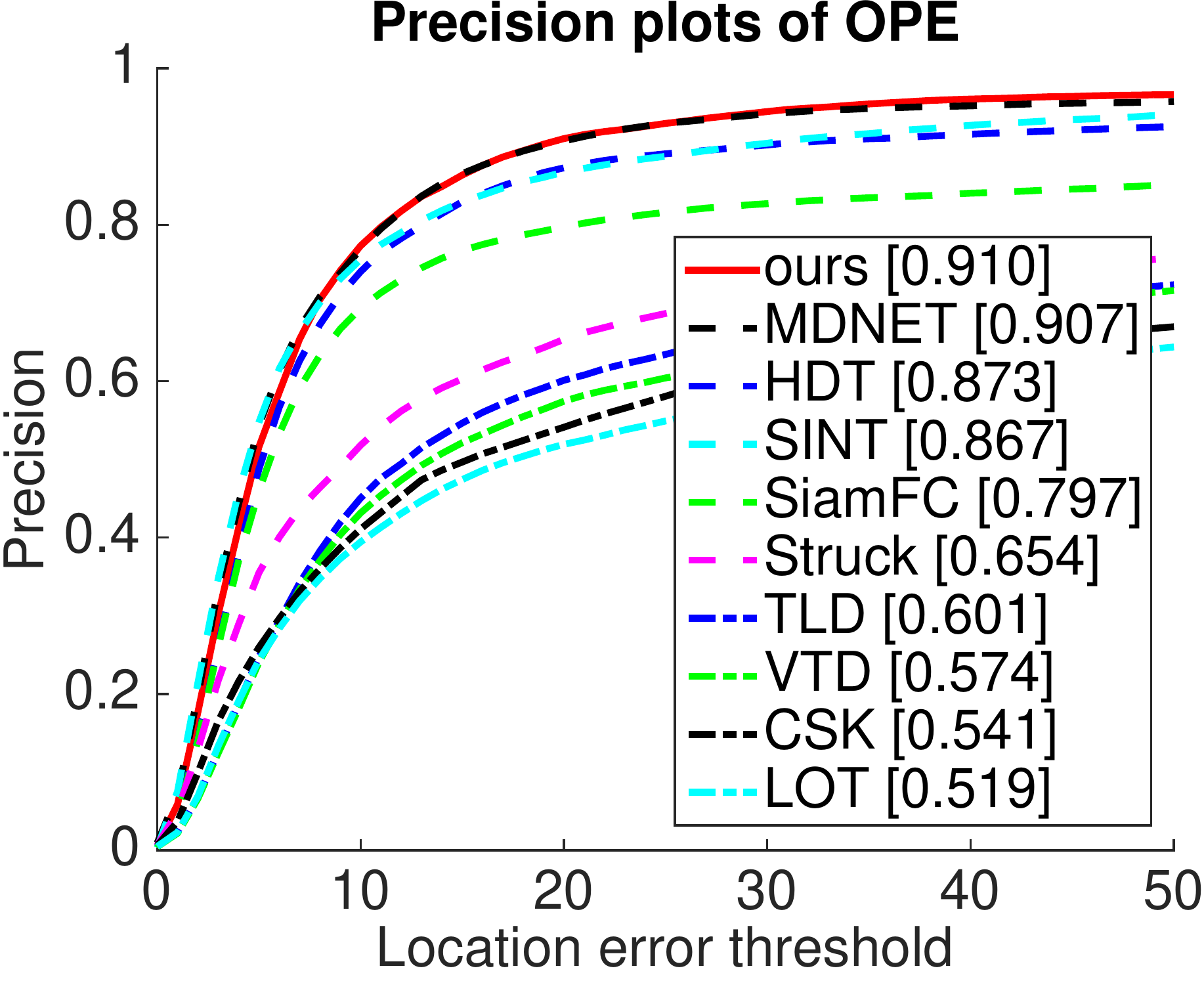}
	\includegraphics[width=0.4\textwidth,height=0.35\textwidth]{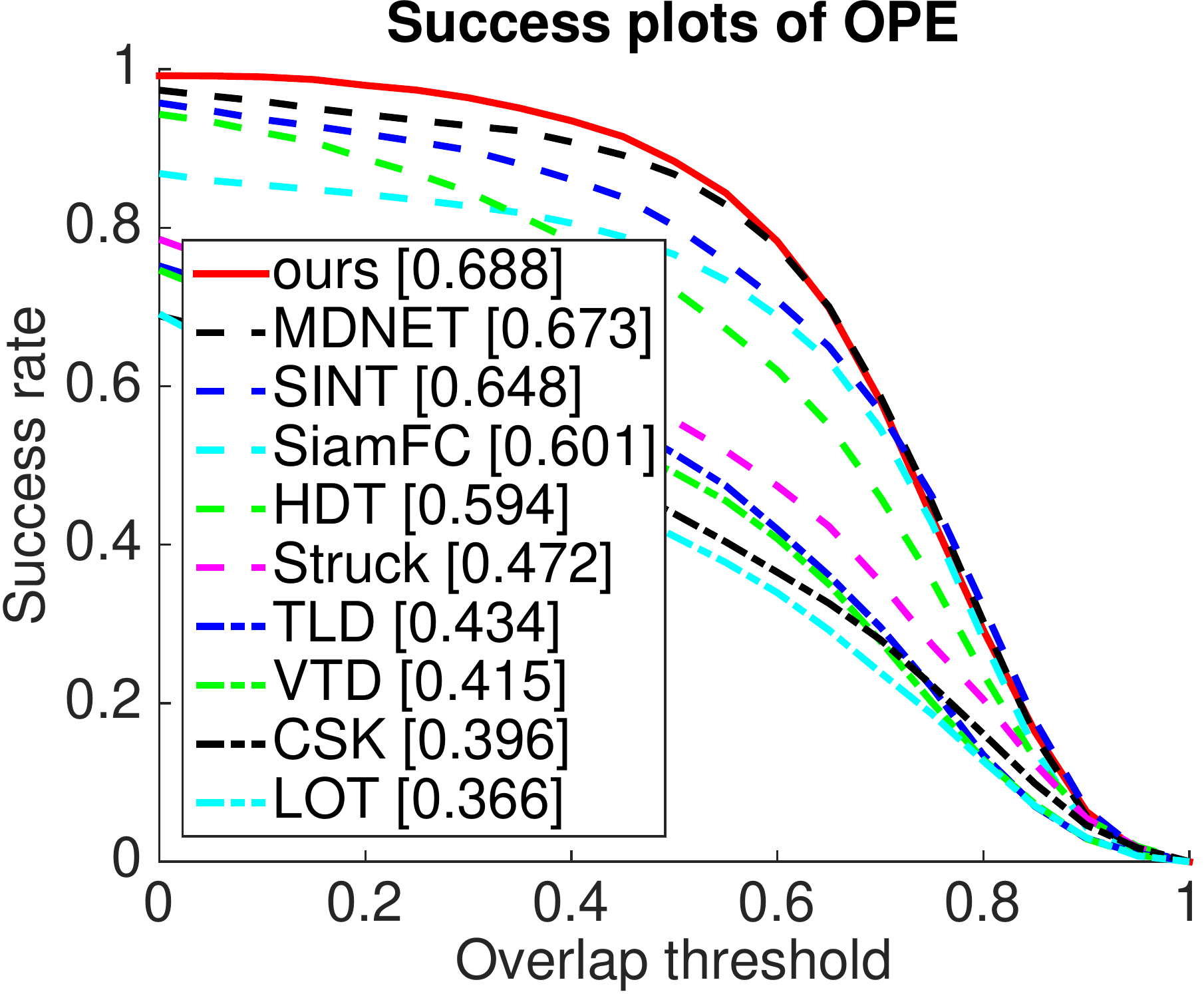}
	\caption{OTB2013 results} 
\end{subfigure}
\begin{subfigure}{0.6\textwidth}
	\includegraphics[width=0.4\textwidth,height=0.35\textwidth]{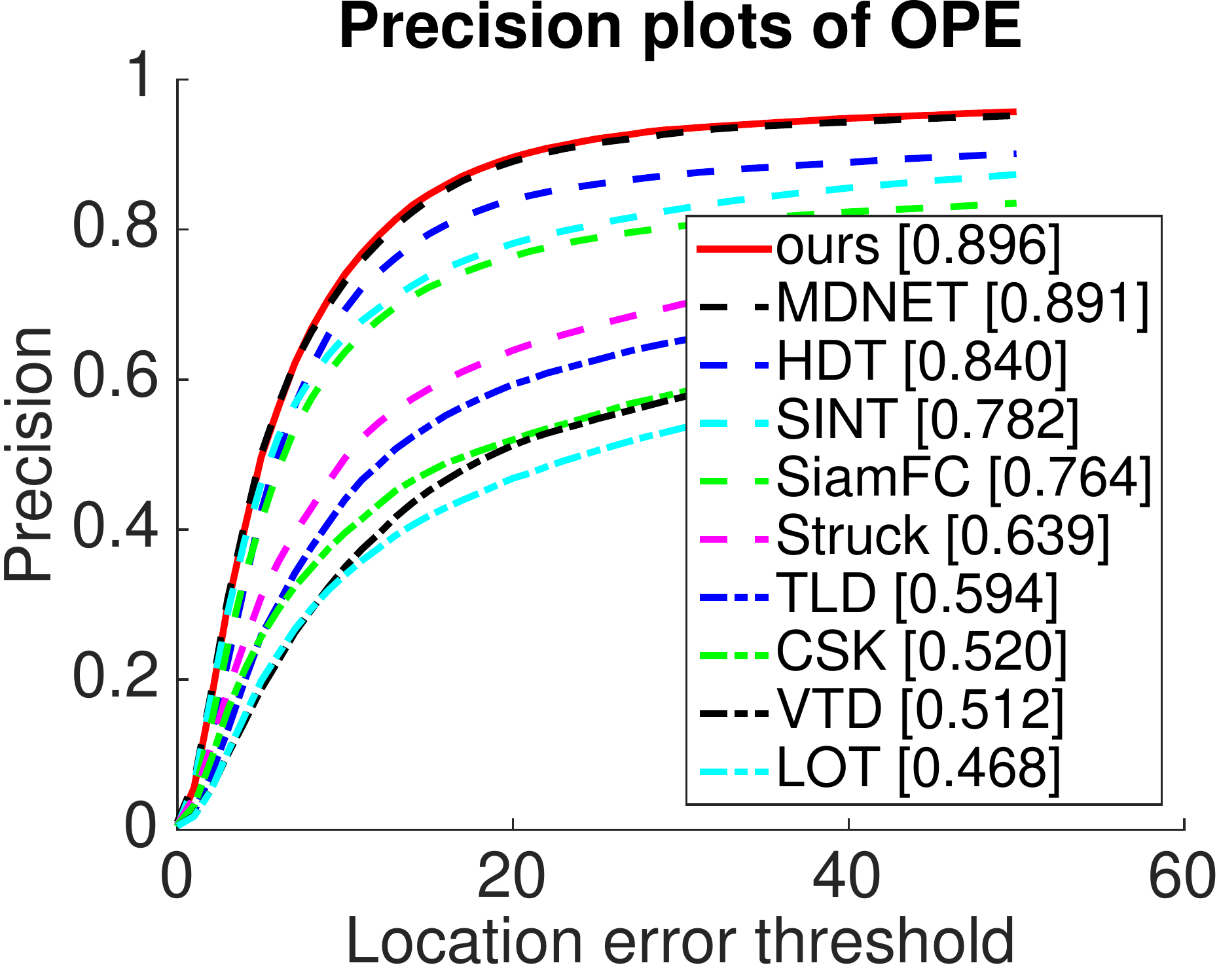}
	\includegraphics[width=0.4\textwidth,height=0.35\textwidth]{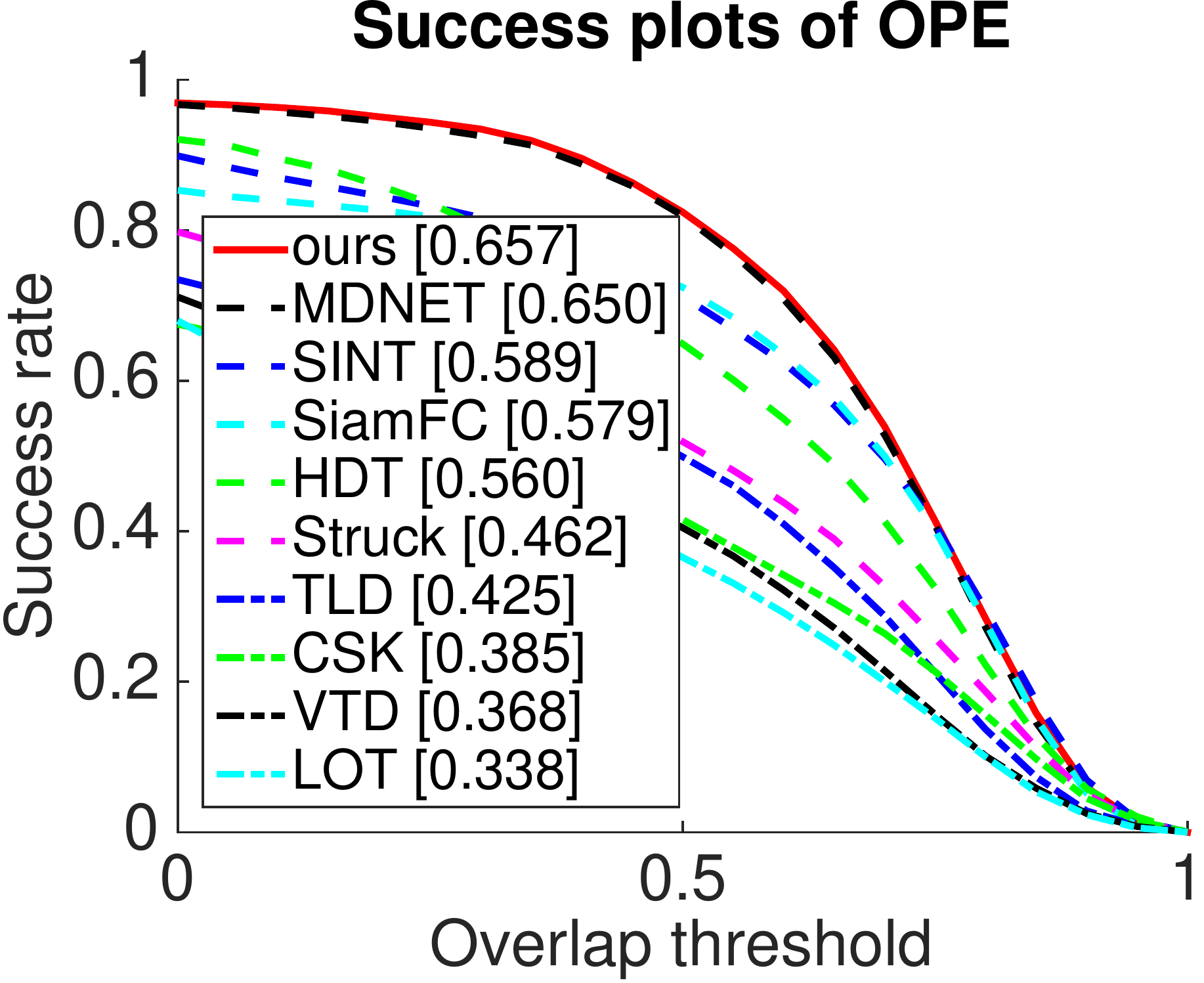}
	\caption{OTB2015 results}
\end{subfigure}
\caption{Precision and success plot on OTB2013\cite{OTB2013} and OTB2015\cite{OTB2015}. The numbers in the legend indicate the representative precision at 20 pixels for precision plots, and the area-under-curve scores for success plots.}
\label{fig:OTB}
\end{figure}

\begin{figure}[!ht]
  \begin{subfigure}{0.235\textwidth}
  \includegraphics[width=1\textwidth,height=.9\textwidth]{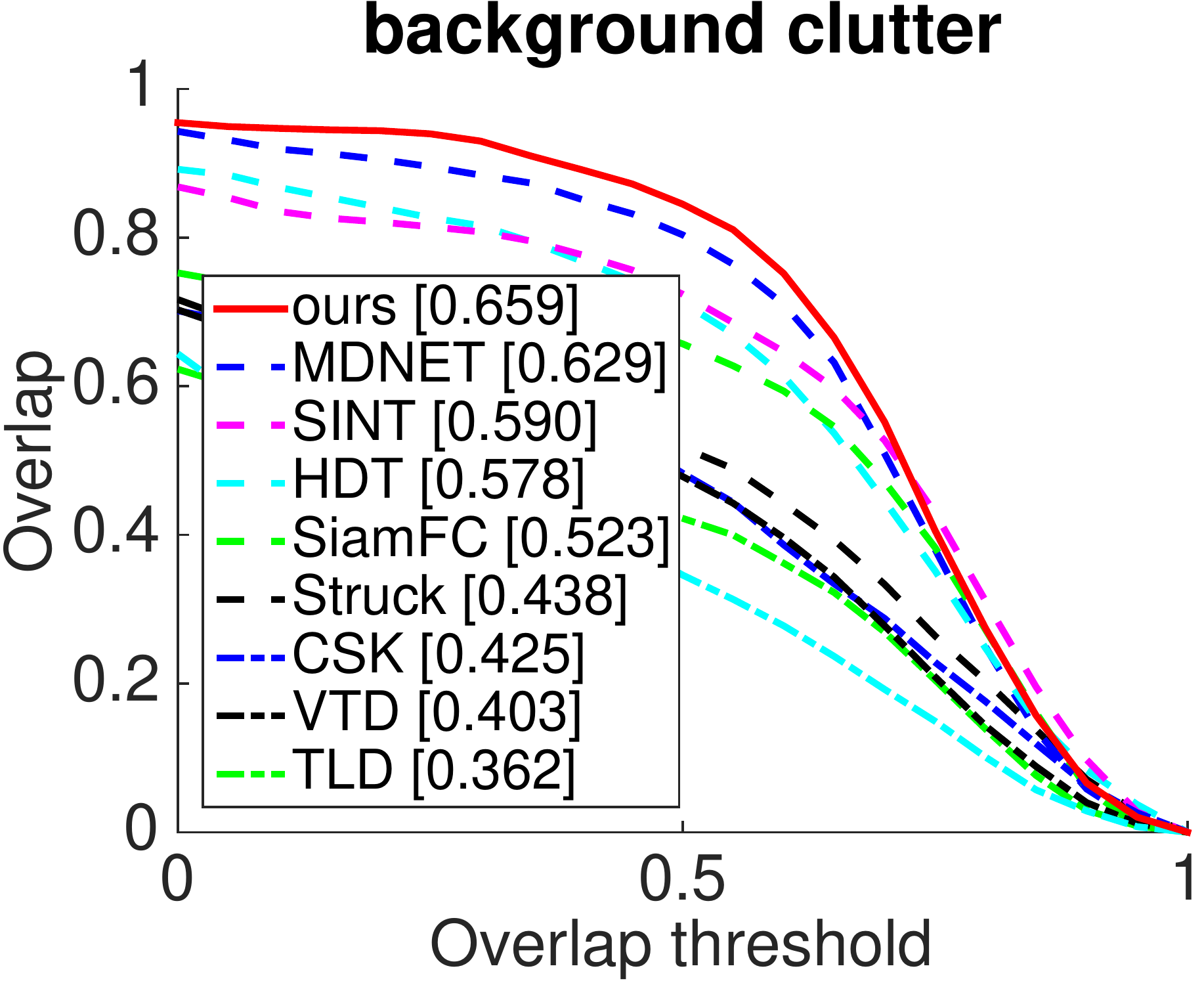}
  \caption{Backgound clutter}
  \end{subfigure}
  \begin{subfigure}{0.235\textwidth}
  \includegraphics[width=1\textwidth,height=.9\textwidth]{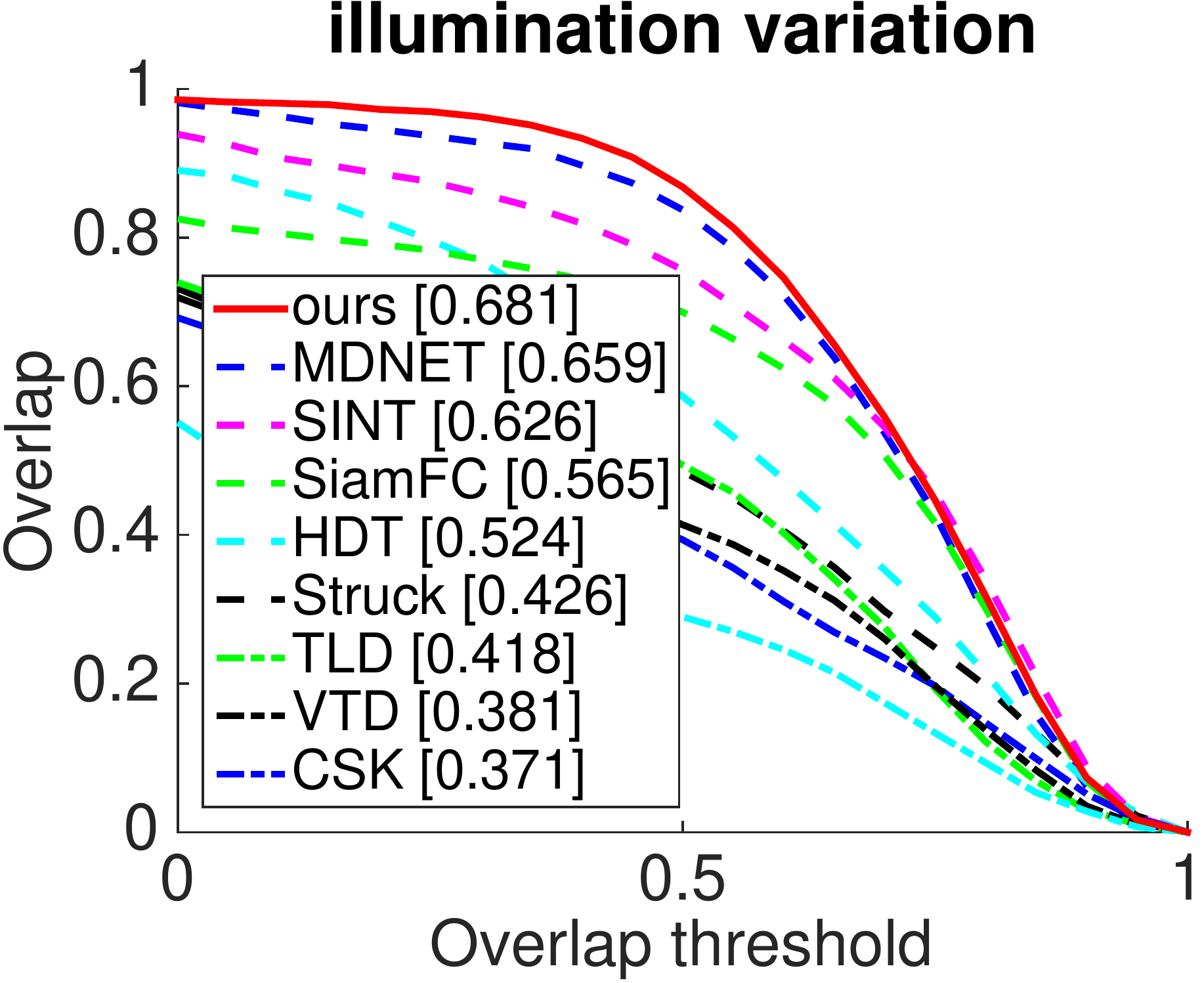}
  \caption{Illumination variation}
  \end{subfigure}
  \begin{subfigure}{0.235\textwidth}
  \includegraphics[width=1\textwidth,height=.9\textwidth]{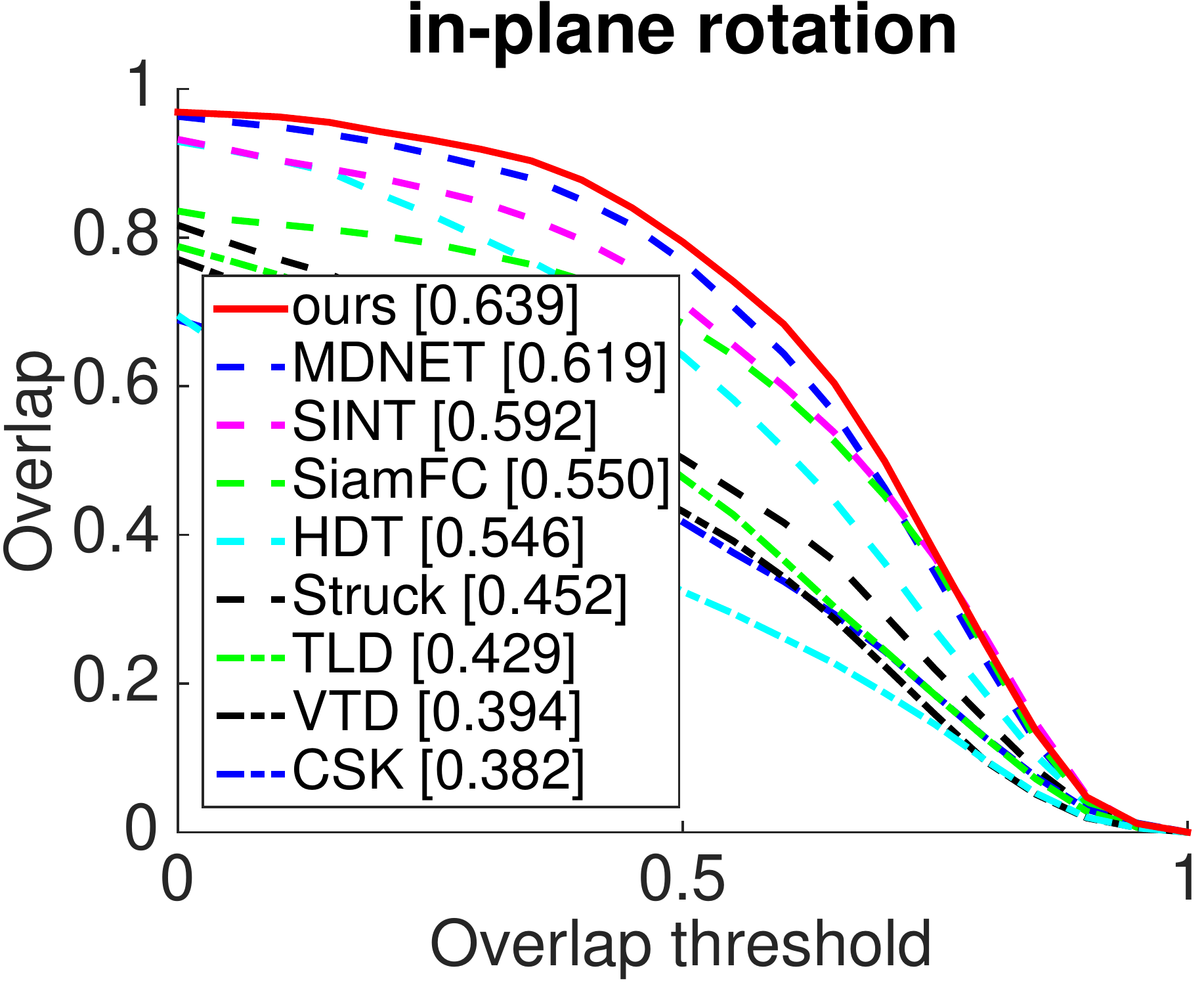}
  \caption{In-plane rotation}
  \end{subfigure}
  \begin{subfigure}{0.235\textwidth}
  \includegraphics[width=1\textwidth,height=.9\textwidth]{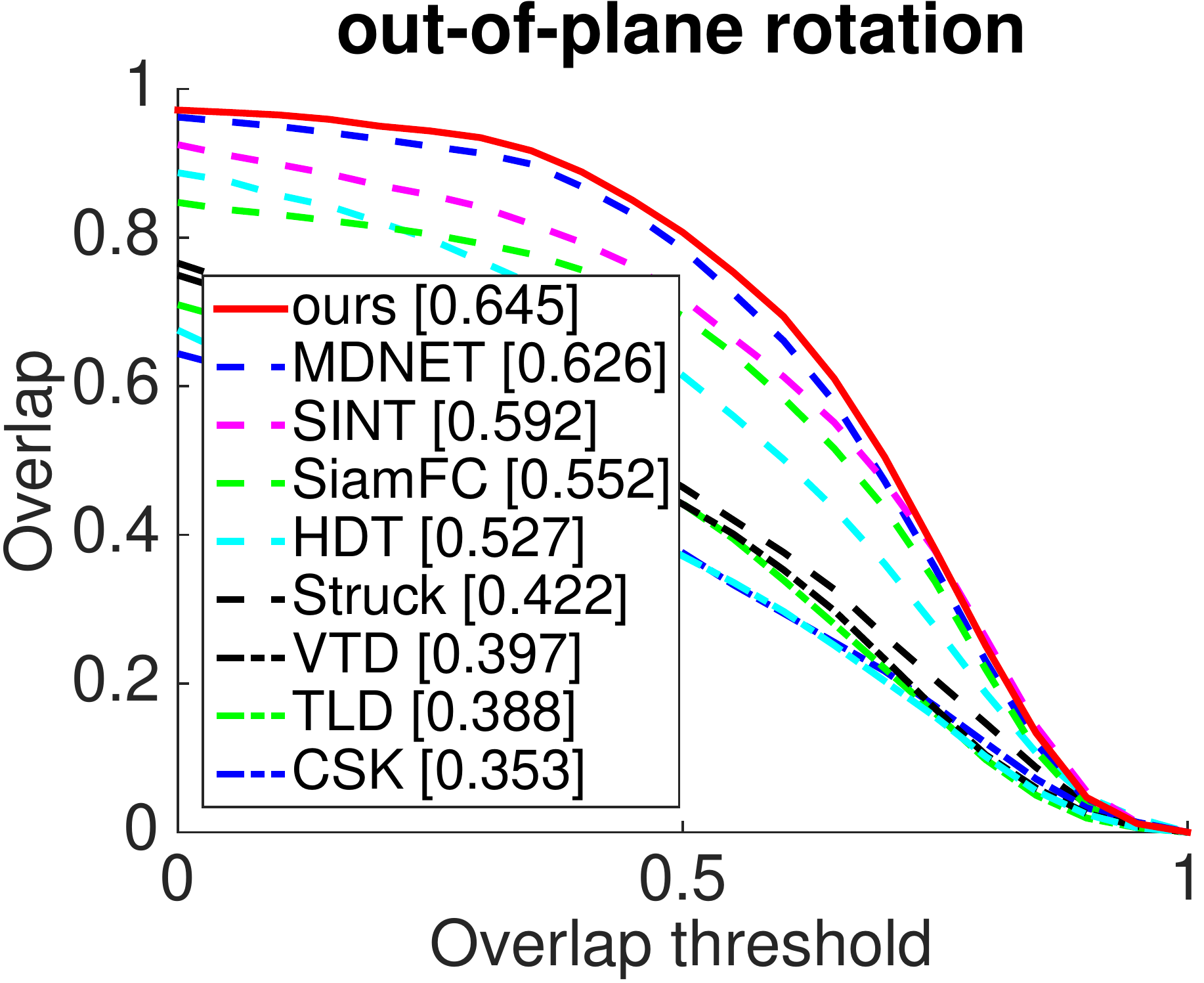}  
  \caption{Out-of-plane rotation}
  \end{subfigure}
\caption{The success plots with four challenge attributes: background clutter, illumination variation, in-plane rotation, and out-of-plane rotation.}
\label{fig:challenges}
\end{figure}

\subsection{Ablation studies}
\label{ssec:subhead}
	In this experiment, ablation studies are employed to illustrate the effectiveness of the adaptive keyframe scheduling algorithm. Fig. \ref{fig:score_threshold_relation} illustrates how the keyframe ratio and overlap success correlate with the score threshold on OTB2015. The above illustration represents when the score threshold gets higher, the ratio of keyframes increases. It can be explained that the score threshold affects the number of keyframes directly. The below illustration shows when the score threshold gets higher, the overlap success ratio first increases, and then converges to that of MDNet\cite{MDNET} in the end. It can be appropriately interpreted that there exists an extreme value for the score threshold in our framework. In other words, DFCNet achieves both model effectiveness and robustness with a proper keyframe ratio. When the percentage of keyframes is low (lower than the proper ratio), DFCNet does not have adequate appearance information, which results in poor performance. However, when the percentage maintains around the proper ratio, DFCNet obtains both sufficient inter-frame information and object appearance features, which leads to high accuracy. At the same time, as the flow computation is less expensive than feature extraction, the model is substantially faster than the baseline model MDNet\cite{MDNET}.

	Table \ref{table:KeyFrameRatio} records the tracking accuracy and speed of the DFCNet with the variation of the score threshold on OTB2015. DFCNet/w indicates a method without an adaptive keyframe mechanism. DFCNet-\textbf{number} represents a standard DFCNet with an adaptive keyframe mechanism, and the score threshold is set to \textbf{number}. It shows the effectiveness of the adaptive keyframe mechanism. 

\begin{table}[!ht]
\small
KFR = KeyFrame Ratio, OS = Overlap Success, SU = SpeedUp\\
\centering
\begin{tabular}{| c | c | c | c | c |}
\hline
Metric & KFR (\%) & OS (\%) & Speed (fps) & SU (\%) \\
\hline
DFCNet/w & 33.3 & 59.0 & 4.06 & 120.9 \\
\hline
DFCNet-2 & 36.8 & 61.2 & 3.77 & 104.1 \\
DFCNet-6 & 44.5 & 62.3 & 3.53 & 91.8 \\
\textbf{DFCNet-10} & \textbf{52.4} & \textbf{65.7} & \textbf{2.95} & \textbf{60.3}  \\
DFCNet-18 & 69.7 & 64.5 & 2.50 & 36.1 \\
DFCNet-26 & 85.0 & 65.7 & 2.21 & 20.2 \\
DFCNet-34 & 96.3 & 65.2 & 1.90 & 3.4 \\
DFCNet-42 & 99.3 & 65.3 & 1.86 & 1.0 \\
DFCNet-50 & 99.9 & 65.1 & 1.84 & 0 \\
DFCNet-58 & 100 & 65.0 & 1.84 & 0 \\
\hline
MDNet & - & 65.0 & 1.84 & - \\
\hline
\end{tabular}
\caption{The table of overlap success and speed with the variation of score threshold on OTB2015 (DFCNet vs. MDNet). The line with bold is the setting we used in the examination on OTB.}
\label{table:KeyFrameRatio}
\end{table}




\subsection{Results on OTB}
\label{ssec:subhead}
	OTB2013\cite{OTB2013} has 50 fully annotated videos with great variation. And OTB2015\cite{OTB2015} extends to 100 video sequences. In this experiment, we compare our method against trackers that published at top conferences and journals, including MDNet\cite{MDNET}, HDT\cite{HDT}, SINT+\cite{SINT+}, SiamFC\cite{SiamFC}, Struck\cite{Struck}, TLD\cite{TLD}, CSK\cite{CSK}, LOT\cite{LOT}, VDT\cite{VDT}.

	Fig. \ref{fig:OTB} illustrates the overlap success and precision plots based on bounding box ratio and center location error, respectively. It clearly shows that DFCNet sightly exceeds MDNet and outperforms the popular trackers. For further performance analyses, we also represent the results on various challenge attributes in OTB2015, such as background clutter, illumination variation, in-plane rotation, and out-of-plane rotation. Fig. \ref{fig:challenges} shows that our tracker effectively handles these challenges while others obtain relatively low scores.

\section{Conclusion}
\label{sec:print}
	In this work, we propose a flow collaborative network that utilizes inter-flow information in consecutive video frames. The algorithm only runs the complex feature network on sparse keyframes and propagates the features to other frames via the optical flow map. Besides, an adaptive keyframe scheduling mechanism is employed to maximize the benefits of both appearance features and temporal information. Our method can realize model effectiveness under the premise of maintaining model robustness. The approach is validated on benchmarks OTB2013\cite{OTB2013} and OTB2015\cite{OTB2015}. It is around 60\% faster than MDNet\cite{MDNET} on OTB2015, which indicates the effectiveness of our method. Moreover, DFCNet performs favorably against existing popular trackers in accuracy and significantly advances the practice of visual tracking tasks.

\vfill\pagebreak
\clearpage

\label{sec:refs}

\bibliographystyle{IEEEbib}
\bibliography{ms}

\end{document}